\documentclass[10pt,twocolumn,letterpaper]{article}

\usepackage{cvpr}              

\usepackage[accsupp]{axessibility}
\usepackage{graphicx}
\usepackage{amsmath}
\usepackage{amssymb}
\usepackage{booktabs}

\usepackage{times}
\usepackage{epsfig}
\usepackage{tikz}
\usepackage{comment}
\usepackage{color}
\usepackage{xcolor}
\usepackage{amsfonts}
\usepackage[hang, flushmargin]{footmisc}         
\usepackage[inline]{enumitem}
\usepackage{algorithm}
\usepackage{algpseudocode}
\usepackage{multirow}
\usepackage{cite}
\usepackage{balance}
\usepackage{adjustbox}

\usepackage[pagebackref,breaklinks,colorlinks]{hyperref}

\usepackage[capitalize]{cleveref}
\crefname{section}{Sec.}{Secs.}
\Crefname{section}{Section}{Sections}
\Crefname{table}{Table}{Tables}
\crefname{table}{Tab.}{Tabs.}

\def\cvprPaperID{26} 
\def\confName{CVPR}
\def\confYear{2023}

\begin{document}

\title{Revisiting Class Imbalance for End-to-end Semi-Supervised Object Detection}

\author{Purbayan Kar, Vishal Chudasama, Naoyuki Onoe, Pankaj Wasnik\thanks{Pankaj Wasnik is the corresponding author.} \\ Media Analysis Group, Sony Research India, Bangalore, India\\
\tt\normalsize\{purbayan.kar, vishal.chudasama1, naoyuki.onoe, pankaj.wasnik\}@sony.com}

\maketitle

\begin{abstract}
Semi-supervised object detection (SSOD) has made significant progress with the development of pseudo-label-based end-to-end methods. However, many of these methods face challenges due to class imbalance, which hinders the effectiveness of the pseudo-label generator. Furthermore, in the literature, it has been observed that low-quality pseudo-labels severely limit the performance of SSOD. 
In this paper, we examine the root causes of low-quality pseudo-labels and present novel learning mechanisms to improve the label generation quality. To cope with high false-negative and low precision rates, we introduce an adaptive thresholding mechanism that helps the proposed network to filter out optimal bounding boxes. We further introduce a Jitter-Bagging module to provide accurate information on localization to help refine the bounding boxes. Additionally, two new losses are introduced using the background and foreground scores predicted by the teacher and student networks to improvise the pseudo-label recall rate. Furthermore, our method applies strict supervision to the teacher network by feeding strong \& weak augmented data to generate robust pseudo-labels so that it can detect small and complex objects. Finally, the extensive experiments show that the proposed network outperforms state-of-the-art methods on MS-COCO and Pascal VOC datasets and allows the baseline network to achieve 100\% supervised performance with much less (i.e., 20\%) labeled data.
\end{abstract}

\section{Introduction}
\label{sec:intro}
\begin{figure}[t!]
    \centering
        \subfloat[]{\includegraphics[width=0.48\columnwidth, height = 0.11\textheight]{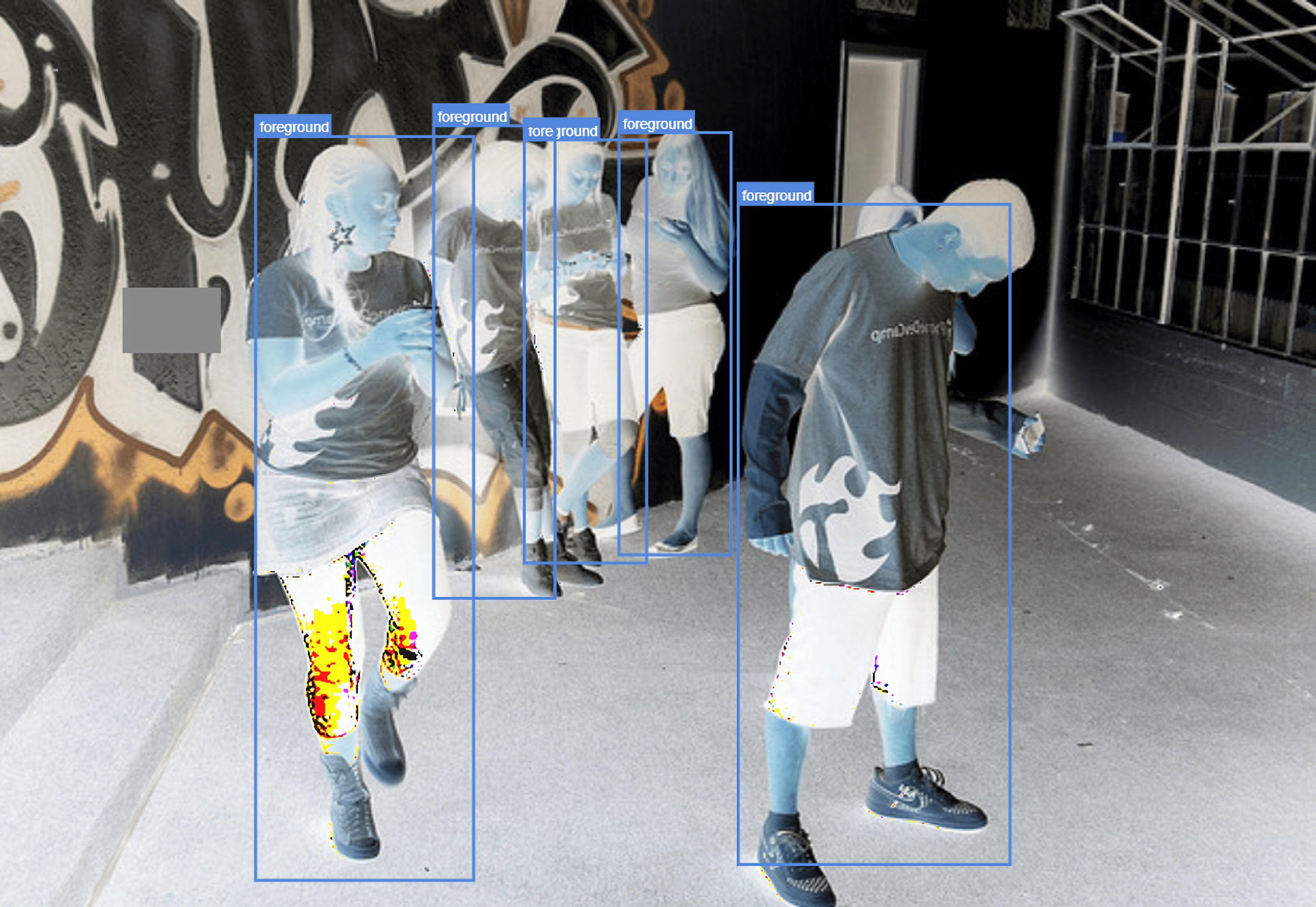}} \hspace{0em}
        \subfloat[]{\includegraphics[width=0.48\columnwidth, height = 0.11\textheight]{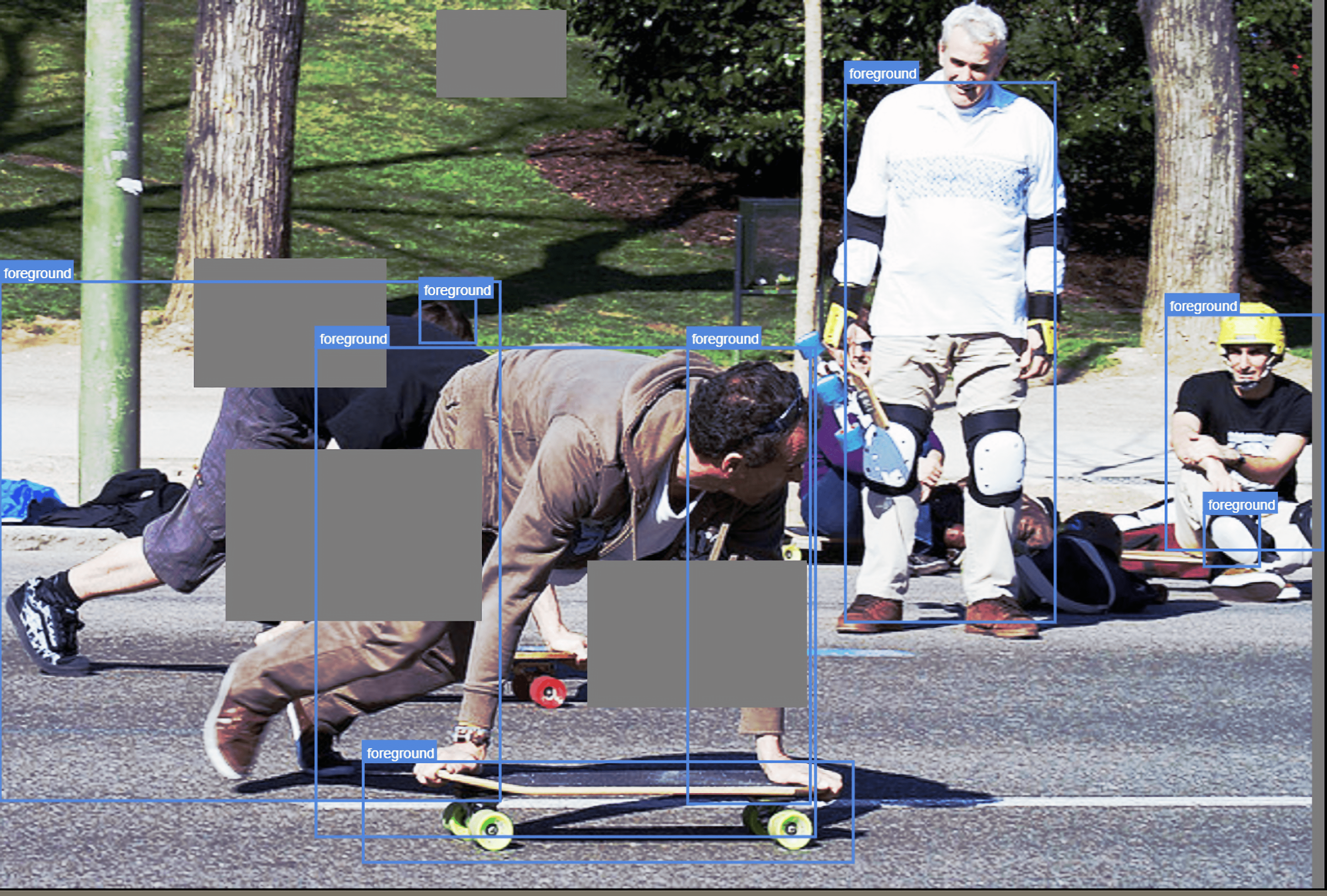}} \vspace{-1em}
        \subfloat[]{\includegraphics[width=0.48\columnwidth, height = 0.111\textheight]{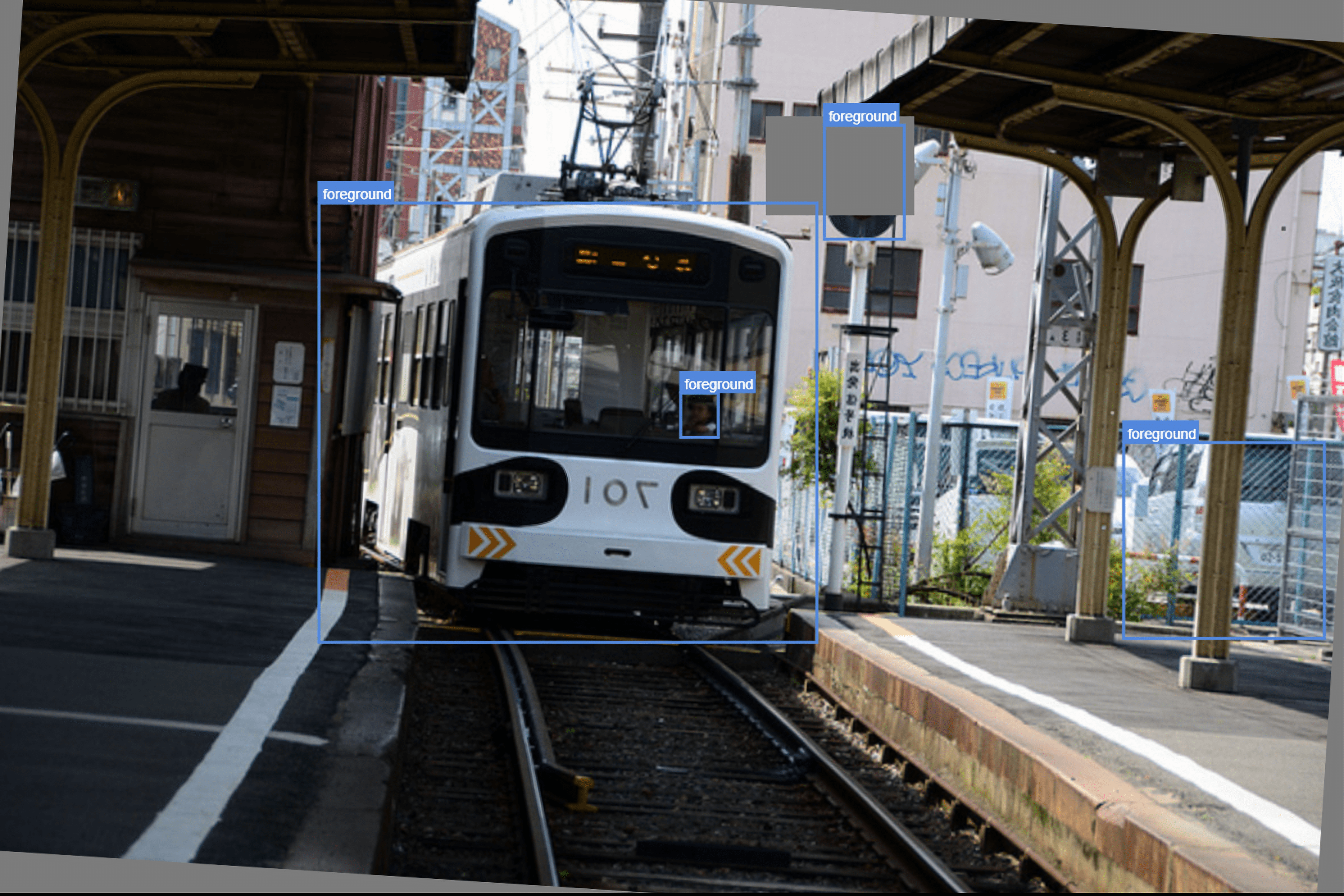}} \hspace{0em}
        \subfloat[]{\includegraphics[width=0.48\columnwidth, height = 0.111\textheight]{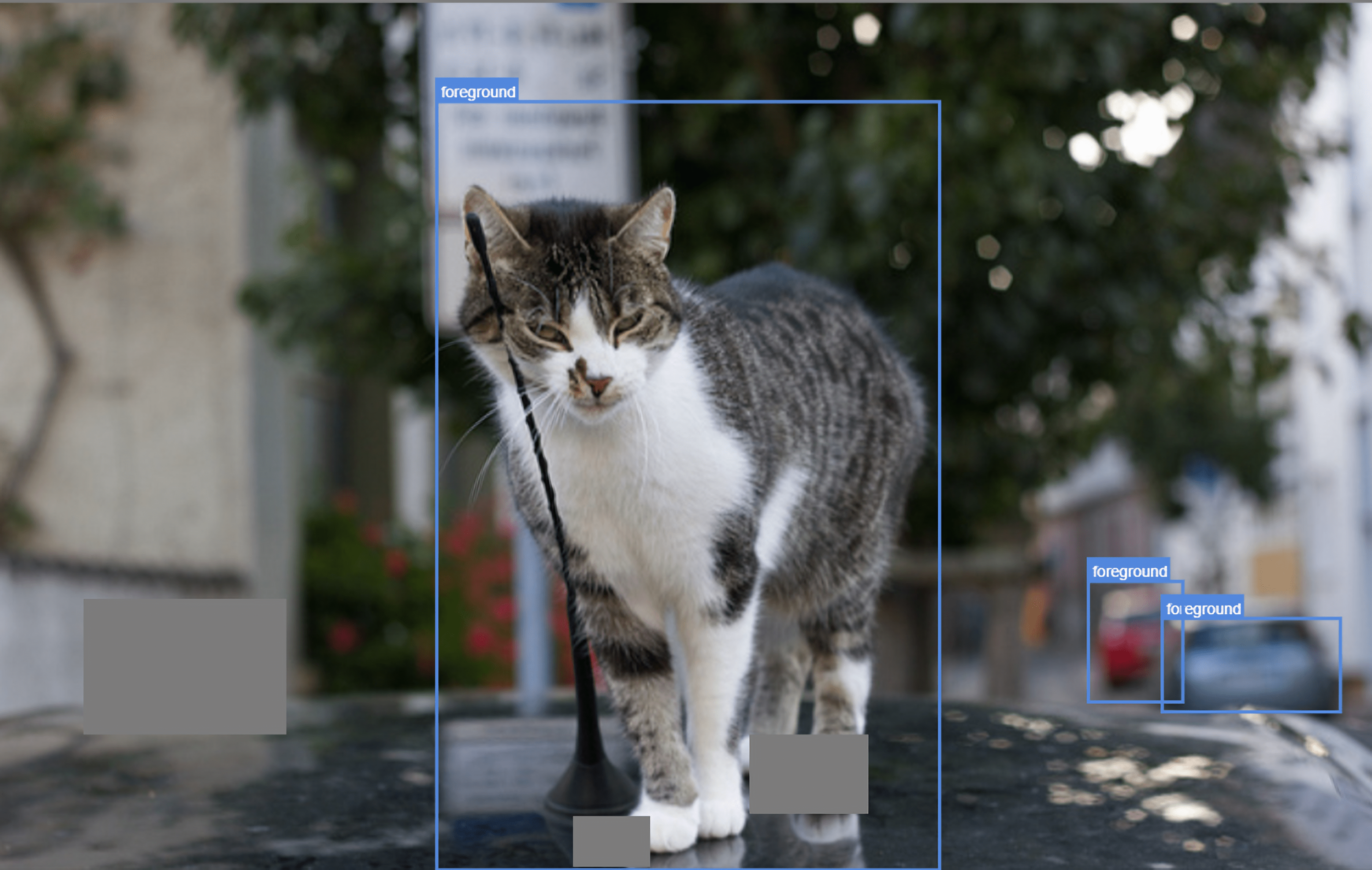}}\vspace{-1em}
\caption{Foreground detection on challenging scenarios such as bad exposure, high perturbations, small, and blur.} \label{fig:res_page_1}
\end{figure}
\vspace{-0.5em}
\begin{figure}[t!]
    \centering
        \includegraphics[width=0.99\columnwidth]{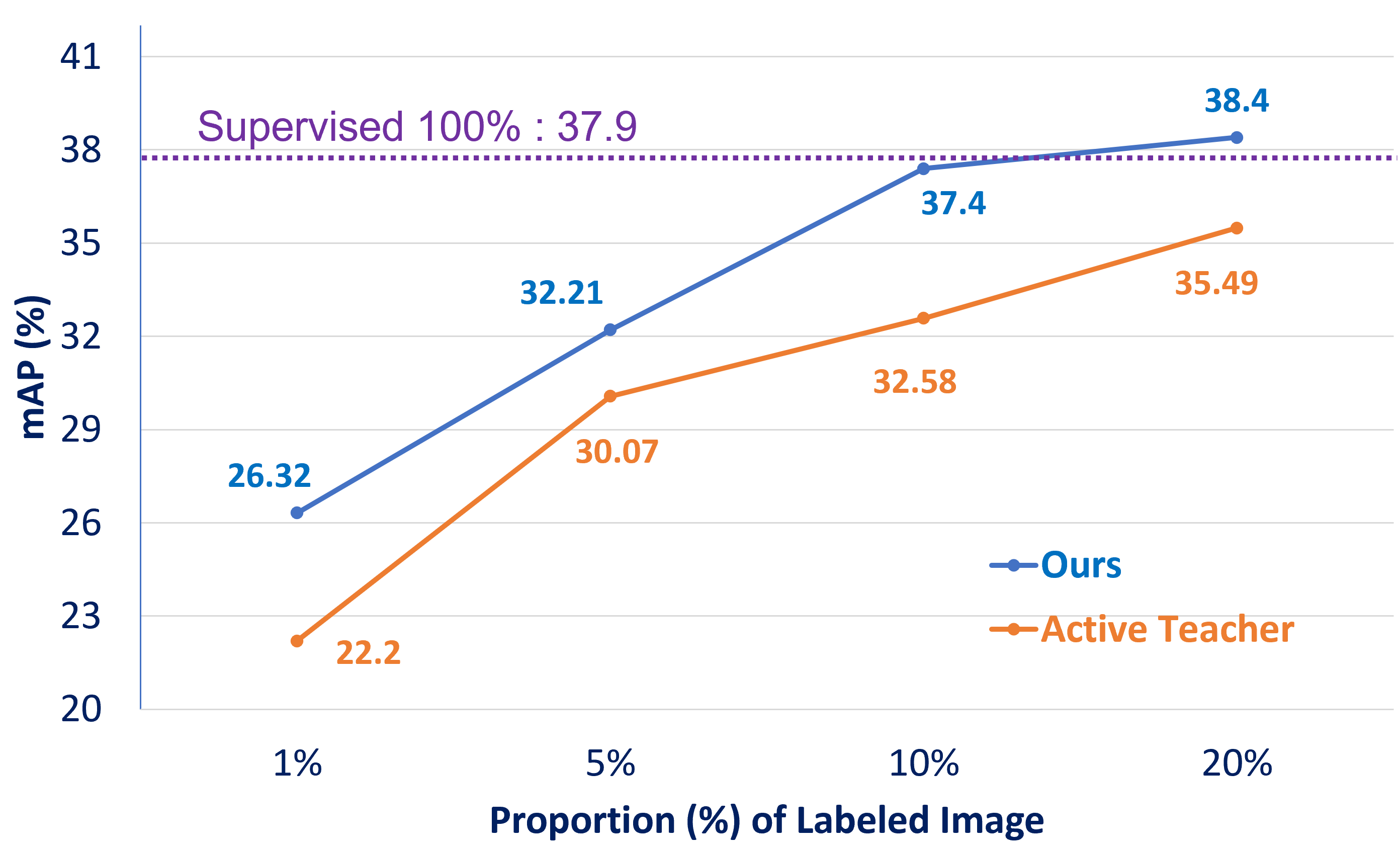}\vspace{-0.5em}
\caption{Comparison between proposed network and the existing SOTA method called Active Teacher \cite{ActiveTeacher} on MS-COCO dataset.} \label{fig:intro_chart}
\end{figure}
The semi-supervised learning (SSL) theory provides a useful illustration of how the vast amount of unlabeled data can be exploited using a small labeled data set \cite{SSL2020survey}. In this work, we revisit the problem of SSL-based object detection (SSOD), in which an object detector is trained with a large amount of unlabeled data and a small amount of labeled bounding boxes. To achieve this, existing SSOD methods typically use two strategies: consistency-based SSOD \cite{tang2021proposal, jeong2019consistency} and pseudo-tagging-based SSOD \cite{xu2021end, ActiveTeacher, PoshingTeacher, DCST, MAGCP, LabelMatch}. Consistency-based approaches train their detector by minimizing the inconsistency between the predictions of unlabeled data with different perturbations. Their performance is highly dependent on the design of the perturbations and the consistency measurement.
Recently, pseudo-labeling-based frameworks \cite{xu2021end, ActiveTeacher, PoshingTeacher, DCST, MAGCP, LabelMatch} have become popular. These methods follow teacher-student scheme in which the teacher network generates pseudo-labels using unlabeled data. Concurrently, the student network is trained at each iteration with the predicted pseudo-labels and few labeled data. 
The benefit of such learning is that as the network converges during training, the quality of the pseudo-labels increases. However, a high-quality pseudo-label requires both precise classification and localization \cite{DCST}. In this paper, we examine the root cause of the negative impact of low-quality pseudo labels.

Low-quality pseudo-labeling can aggravate class imbalance issues, resulting in a high false-negative rate (i.e., failing to identify objects from the less prevalent classes) and a low precision rate (i.e., incorrectly identifying objects from the less common classes). To address this, the existing pseudo-label-based SSOD frameworks \cite{tang2021humble, zhou2021instant, xu2021end} use a hand-crafted threshold to distinguish pseudo-bounding boxes for student training. However, the hard threshold as a hyperparameter must be carefully tuned and dynamically adjusted according to the model capability in different time steps \cite{consistent-teaching}. To carefully discriminate pseudo-bounding boxes, we introduce an adaptive threshold filter mechanism that adjusts the threshold based on background/foreground bounding boxes at each time step. From experimental analysis, we also proved the effectiveness of adaptive threshold mechanism over static hand-crafted threshold.

Applying pseudo-labels directly to object detection raises the problem of imprecise bounding box localization \cite{DCST}. Xu \etal \cite{xu2021end} also analyzed this effect and found 
that bounding boxes with high foreground scores may not provide accurate localization information and therefore not suitable for box regression tasks. To address this issue, we introduce a Jitter-Bagging module to estimate the reliable bounding boxes by measuring the consistency of its regression prediction. 
The effectiveness of the proposed Jitter-Bagging is also validated in the experimental section.

Another problem is the low recall rate of pseudo-labels, which impairs model training and causes many candidate boxes to be mistaken for a background category due to poor matching of pseudo-labels \cite{DCST}. To address this issue, we're introducing two new losses that will help to improve foreground classification accuracy. The first loss is the background similarity loss, which helps the network to match the teacher-generated pseudo-boxes and the predicted boxes. Minimizing this loss ensures that the pseudo-boxes generated by the teacher and predicted by the students become as similar as possible. Another loss function is the foreground-background dissimilarity loss to separate the foreground bounding box from the background boxes. These losses help the proposed network to improve the pseudo-label recall rate, so that it can improve detection performance. 
In addition, we analyzed impact of the Exponential Moving Average (EMA) update mechanism and found that it faces a lag issue that limits its performance in sudden weight fluctuations. To address this, we employed Double EMA (DEMA) \cite{DEMA2}, which gives more weight to the latest observations and removes the lag compared to EMA. As per our knowledge, the DEMA is being used in the SSOD weight update mechanism for the first time, and its effectiveness in detail is discussed in the Section \ref{sec:abl}. Our scheme also applies strict supervision to the teacher network and feeds strong and weak augmented data into the teacher network to generate accurate pseudo-labels. Interestingly, the proposed network can accurately detect the foreground bounding boxes even for small and highly complex objects. This can be verified using Figure \ref{fig:res_page_1}, where some results generated by the proposed network on challenging scenarios are visualized.

We perform extensive experiments with benchmark datasets, namely MS-COCO \cite{lin2014microsoft} and Pascal VOC \cite{pascal} to validate the proposed method. The experimental analysis not only confirms the significant performance gain over SOTA methods, but also shows that the proposed method allows the baseline network to achieve 100\% supervised performance with much less (i.e., 20\%) annotated images in the MS-COCO dataset as shown in Figure \ref{fig:intro_chart}.

Finally, the key contributions of the paper can be summarized as follows:
\begin{itemize}
    \item The existing class imbalance can significantly hinder the efficacy of pseudo-label generators. This issue can be alleviated by incorporating suitable learning mechanisms in end-to-end manner.
    \vspace{-0.7em}
    \item The high false negative and low precision rates can be improvised by carefully distinguishing the pseudo-bounding boxes. To handle this, we propose an adaptive thresholding mechanism that adjusts the threshold based on background/foreground bounding boxes and helps to filter out optimal bounding boxes.
    \vspace{-0.7em}
    \item To provide accurate localization information, we introduce a Jitter-Bagging module for the regression task that helps the proposed network to refine optimal bounding boxes.
    \vspace{-0.7em}
    \item To improve pseudo-label recall rate and detect blurry and distorted small objects as foreground objects, we introduce two new losses: background similarity loss and foreground-background dissimilarity loss.
\end{itemize}

\begin{figure*}[t!]
    \centering
    \includegraphics[width = 0.99\textwidth] {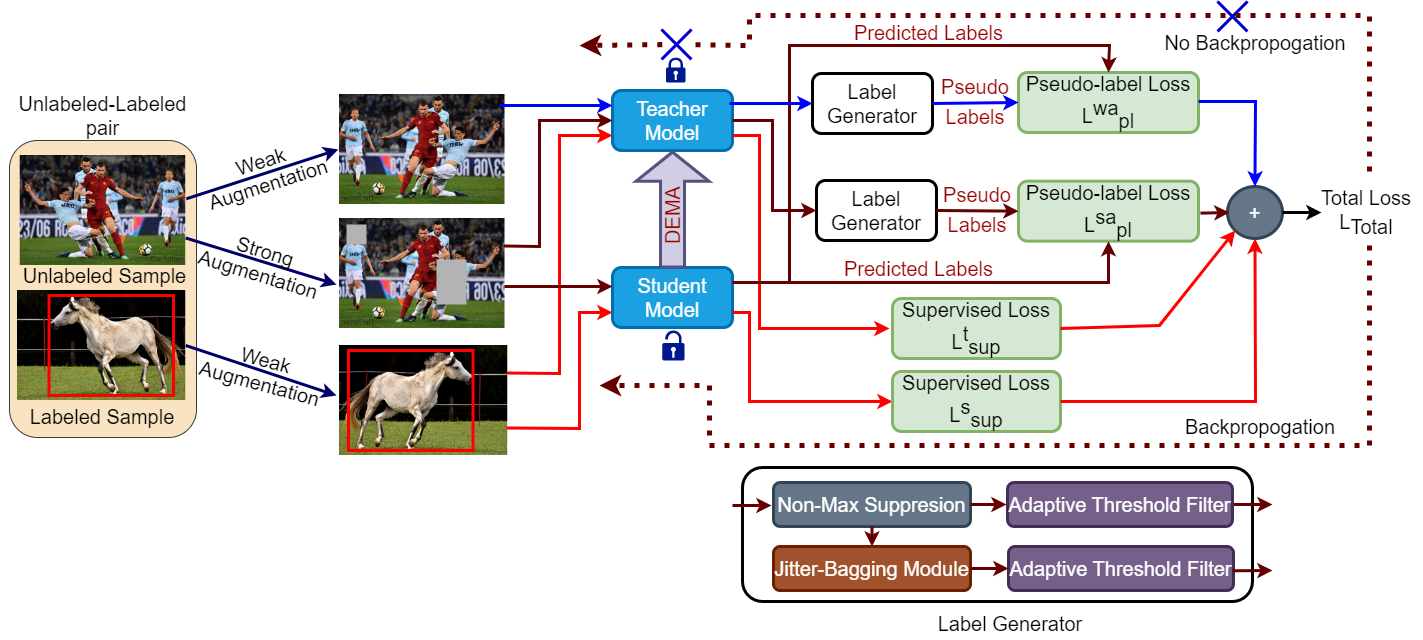}
    \caption{Network architecture of the proposed teacher-student-based network.}
    \label{fig:model}
\end{figure*}
\section{Related works}

Existing SSOD frameworks can be categorized as the pseudo-label-based methods \cite{radosavovic2018data, li2020improving, sohn2020simple, wang2018towards, zoph2020rethinking, xu2021end, zhang2021semi, li2021rethinking, DDT, CST, DTG-SSOD, ActiveTeacher, MAGCP, DCST, PseCo, LabelMatch, PoshingTeacher} and consistency-based methods \cite{tang2021proposal, jeong2019consistency}. The pseudo-label-based works proved better performance against consistency-based SSOD approaches. 

In \cite{sohn2020simple}, Shon \etal introduce a pseudo-labeling-based method but it lacks consideration of serious data imbalance issues. Zhang \etal \cite{zhang2021semi} proposed an adaptive self-training model for class-rebalancing but it requires additional memory module. 
Yang \etal \cite{yang2021interactive} proposed interactive form of self-training to tackle the discrepancies in results. However, their model requires two ROI heads to mine complementary information.
In contrast, Xu \etal \cite{xu2021end} present an end-to-end based soft teacher mechanism to generate better pseudo-labels. Similarly, Tang \etal \cite{tang2021humble} follow the teacher-student dual model framework to generate more consistent pseudo-labels. 
These methods \cite{xu2021end, tang2021humble} perform better than multi-stage based approaches with less complexity, but they still suffers from class imbalance problem. 
Zheng \etal \cite{DDT} observed the effects of single threshold and introduced a two-stage threshold mechanism based dual decoupling framework. Recently, Liu \etal \cite{CST} proposed a cycle self-training network to overcome the coupling effect of teacher-student learning. In \cite{DTG-SSOD}, Li \etal proposed a method that uses the dense guidance teacher directly to monitor student training. Recently, Mi \etal \cite{ActiveTeacher} examined teacher-student learning from the perspective of data initialization where the label set is partially initialized and gradually augmented by evaluating key factors of unlabeled examples. In \cite{DCST}, Wang \etal identified the inconsistency in object proposals and proposed a framework to overcome the harm caused by insufficient quality of pseudo-labels.

Recent, authors in \cite{Unbiasedv2} showed the generalization of the SSOD method to anchor-free detectors and also introduced Listen2Student mechanism to prevent misleading pseudo-labels. Chen \etal \cite{DSL} also studied anchor-free detectors and proposed a dense learning-based framework to generate stable and precise pseudo-labels. In \cite{DenseTeacher}, Zhou \etal proposed replacing sparse pseudo-boxes with dense predictions to obtain rich pseudo-label information. Li \etal \cite{PseCo} introduced noisy pseudo box learning and multi-view scale-invariant learning to provide better pseudo-labels. To take full advantage of labeled data, Li \etal \cite{MAGCP} proposed a multi-instance alignment model that improves prediction consistency based on global class prototypes. Che \etal \cite{LabelMatch} proposed a framework from two perspectives, i.e., distribution-level and instance-level to handle the class imbalance issue. Recently, Zhang \etal \cite{PoshingTeacher} introduced a dual pseudo-label polishing framework to reduce the deviation of pseudo-labels from ground truth through dual-polishing learning.

To address the crucial class imbalance issue and produce better pseudo-labels, we also employ pseudo-label-based teacher-student scheme
and introduce two crucial modules, two new classification losses and a new learning mechanism. 

\section{Methodology}
\subsection{Problem Statement}
This paper aims to perform robust pseudo-label-based end-to-end SSOD where a set of labeled images $D_l = {\{x_{l,i}; y_{l,i}\}}^{N_l}_{i=1}$ 
and a set of unlabeled images $D_u = {\{x_{u,j}\}}^{N_u}_{j=1}$ are used for training. Here, $N_l$ and $N_u$ are the number of labeled and unlabeled images. Further, $x_l$ and $y_l$ denote the image and its ground-truth annotations i.e., class labels and bounding box coordinates, respectively.

\subsection{Overview of the proposed network}
The architectural pipeline of proposed end-to-end network is illustrated in Figure \ref{fig:model}. 
In each training iteration, labeled and unlabeled images are arbitrarily sampled to form an input batch. The teacher network produces the pseudo labels based on the weak and strong augmented unlabeled images. While the student network is trained using weakly augmented labeled images with ground-truth and strongly augmented unlabeled images with pseudo-labels as ground-truth. Here, the student network is trained using the weighted combination of the supervised and pseudo-label loss. This can be expressed mathematically as
\begin{equation}\label{eq:loss}
    L_{Total} = L_{sup} + \lambda \cdot (L^{wa}_{pl} + L^{sa}_{pl}).
\end{equation}
Here, $\lambda$ controls the contribution of pseudo label loss. $L_{sup}$ is the supervised loss consisting $L^t_{sup}$ and $L^s_{sup}$ loss associated with the teacher and student network, respectively. While $L^{wa}_{pl}$ and $L^{sa}_{pl}$ are the pseudo-label losses based on the weak and strong augmented samples, respectively. These losses are mathematically described as
\begin{equation} 
L_{sup} = \frac{1}{N_l}\sum_{i=1}^{N_l}\Big(L^{cls}_{sup}(I_{l,i})+L^{reg}_{sup}(I_{l,i})\Big),
\end{equation}\vspace{-1em}
\begin{equation}
L^{wa}_{pl} = \frac{1}{N_u}\sum_{i=1}^{N_u}\Big(L^{cls}_{pl}(I_{u,i}^{wa})+L^{reg}_{pl}(I_{u,i}^{wa})\Big),
\end{equation}\vspace{-0.5em}
\begin{equation}
L^{sa}_{pl} = \frac{1}{N_u}\sum_{i=1}^{N_u}\Big(L^{cls}_{pl}(I_{u,i}^{sa})+L^{reg}_{pl}(I_{u,i}^{sa})\Big).
\end{equation}
Here, $I_{l,i}$ denotes the $i^{th}$ labeled image. $I_{u,i}^{sa}$ and $I_{u,i}^{wa}$ indicates the $i^{th}$ strong augmented and weak augmented unlabeled image, respectively. $L^{cls}_{sup}$ and $L^{reg}_{sup}$ are the supervised classification and regression loss. Similarly, $L^{cls}_{pl}$ and $L^{reg}_{pl}$ are pseudo-label based classification and regression losses. Here, the number of labeled images and unlabeled images are noted as $N_l$ and $N_u$, respectively.

During the training process, the student network is trained using the weighted loss function given in Eq.\ref{eq:loss}, and the teacher network is updated via the Double Exponential Moving Average (DEMA) update mechanism \cite{DEMA2}. The teacher network predicts many bounding boxes for an unlabeled image. Hence, we employ the Non-Max Suppression (NMS) to eliminate redundancy. Although most of the redundant boxes are removed, some non-foreground candidates may remain. Therefore, only candidates with a foreground score\footnote{The foreground score is defined as the maximum probability of all non-background categories.} greater than an adaptive threshold are retained as pseudo boxes. 
These pseudo boxes are then utilized in the classification loss. To learn box regression, the bounding boxes are passed through our Jitter-Bagging module to select reliable pseudo boxes, which are subsequently refined by the adaptive threshold filter. 

In the following subsections, we discuss the adaptive threshold filter, efficient classification loss, Jitter-Bagging module and update mechanism in detail.


\subsection{Adaptive threshold filter}
The performance of the detector network depends on quality of the pseudo-label. However, quality degrades due to the class imbalance, especially when there are few annotations. For underrepresented classes, the teacher network produces relatively lower confidence score \cite{dave2021evaluating}, which barely survives the large threshold $\tau$. On the other hand, simply lowering $\tau$ leads to noisier pseudo-labels in common classes. Therefore, we propose adaptive threshold filter that adjusts the threshold value based on the confidence scores of the background and foreground bounding boxes for each category.
The adaptive threshold (i.e., $\tau_a$) is mathematically defined for $N_b^{fg}$ number of foreground and $N_b^{bg}$ number of background bounding boxes as
\begin{equation}
    \tau_{a} = \Bigg\lfloor\Bigg(\frac{\frac{1}{N_{b}^{fg}}\sum_{i=1}^{N_{b}^{fg}}\mathcal S_{i}^{fg}}{\frac{1}{N_{b}^{bg}}\sum_{j=1}^{N_{b}^{bg}}\mathcal S_{j}^{bg}}\Bigg)^{\gamma}\Bigg\rfloor,
\end{equation}
where $\gamma$ controls the degree of the underrepresented classes and it is set to 0.05, $\lfloor\cdot\rfloor$ indicates the closest decimal floor function for single precision (e.g. 0.94 will be set to 0.9). Here, $\mathcal S_i^{fg}$ and $\mathcal S_j^{bg}$ denote the scores obtained from the $i^{th}$ and $j^{th}$ foreground and background bounding boxes.  

At the beginning of the training, when parameters of the networks are not fully learned, the majority is mispredictions, i.e., predominantly background predictions than foreground predictions. So adaptive threshold outputs a relatively smaller value and as the training progresses towards convergence, foreground predictions increases and thus, the adaptive threshold provides a larger value and more constraints for selecting appropriate bounding boxes. 

\subsection{Efficient classification loss}
For classification task, the overall loss is obtained by combining four different losses: foreground classification loss (i.e., $L^{cls}_{fg}$), background classification loss (i.e., $L^{cls}_{bg}$), background similarity loss (i.e., $L_{bg}^{sim}$) and foreground-background dissimilarity loss (i.e., $L_{fg-bg}^{dissim}$). The overall classification loss function can be defined as
\begin{equation}
    L_{pl}^{cls} = L^{cls}_{fg} + L^{cls}_{bg} + L_{bg}^{sim} + L_{fg-bg}^{dissim}.
\end{equation}
\textbf{1) Foreground classification loss:} 
Given student-generated foreground bounding boxes (i.e., $b^{fg}$), the foreground classification loss is defined as
\begin{equation}
    L^{cls}_{fg} =\frac{1}{N_b^{fg}}\sum_{i=1}^{N_b^{fg}}l_{cls}(b_{i}^{fg}(s),\mathcal{B}_{cls}), 
\end{equation}
where $\mathcal{B}_{cls}$ denotes the set of teacher-generated pseudo boxes used for classification, $l_{cls}$ is the box classification loss\footnote{We use standard cross-entropy loss function as classification loss.}, $N_b^{fg}$ is the number of box candidates of box set ${b^{fg}}$.\\
\noindent\textbf{2) Background classification loss:} We employ the same loss function proposed by Xu \etal \cite{xu2021end} for the background classification loss. Given background bounding boxes (i.e., ${b^{bg}}$), the classification loss is calculated as
\begin{equation}
    L^{cls}_{bg} = \sum_{j=1}^{N_b^{bg}}\delta_{j}l_{cls}(b_{j}^{bg}(s),\mathcal{B}_{cls}) 
\end{equation}
where, $\delta_j$ denoted as reliability weighting factor associated with $j^{th}$ sample and the same can be expressed as 
\begin{equation}
    \delta_j = \frac{r_j}{\sum_{k=1}^{N_b^{bg}}{r_k}}
\end{equation}
Here $r_j$ is the reliability score for $j^{th}$ background box candidate, $N_b^{bg}$ is the number of box candidates of box set ${b^{bg}}$. 
\\
\textbf{3) Background similarity loss:} Inspired from \cite{riemannian}, we introduce a novel loss to match the scores obtained from the background bounding boxes generated through teacher and student networks. Minimizing this loss will ensure that the teacher-generated pseudo boxes and student-predicted boxes become as similar as possible. The background similarity loss can be expressed as follows:
\begin{equation}\label{eq:similarity}
    L_{bg}^{sim} = \frac{1}{N_{b}^{bg}}\sum_{i=1}^{N_{b}^{bg}}\beta\cdot\log(|e^{|\mathcal S_{i}^{bg}(s)|} - e^{|\mathcal S_{i}^{bg}(t)|}|+1), 
\end{equation}
where, $\beta$ denotes the controlling parameter, $\mathcal S_{i}^{bg}(t)$ and $\mathcal S_{i}^{bg}(s)$ indicates $i^{th}$ scores obtained from the background bounding boxes generated using teacher and student networks, respectively. \\
\textbf{4) Foreground-Background dissimilarity loss:} 
Inspired from relativistic average discriminator \cite{relativistic}, a novel loss is introduced to separate out the foreground and background bounding boxes. The introduced loss considers the dissimilarity between scores obtained from background and foreground bounding boxes. Mathematically, this can be addressed as follows:
\begin{equation}\label{eq:dissimilarity}
   L_{fg-bg}^{dissim} = \frac{1}{N_{b}^{fg}}\sum_{i=1}^{N_{b}^{fg}}\big(1-|\mathcal S_{i}^{fg}(s)-\frac{1}{N_{b}^{bg}}\sum_{j=1}^{N_{b}^{bg}}\mathcal S_{j}^{bg}(s)|\big).
\end{equation}
Here, $\mathcal S_{i}^{fg}(s)$ and $S_{i}^{bg}(s)$ indicate $i^{th}$ and $j^{th}$ scores obtained from the student-generated foreground and background bounding boxes. 

\subsection{Jitter-Bagging module}
Xu \etal \cite{xu2021end} found that the selection of the teacher-generated pseudo boxes according to the foreground score is not suitable for box regression. 
To tackle this issue, we introduce a Jitter-Bagging module where we sample a jittered-box around the teacher-generated pseudo box candidate $b_i$ and then feed to the teacher network to obtain refined box $b_{i}^{'}$. This can be formulated as
\begin{equation}
    b_{i}^{'} = f_{Jitter}(b_i).
\end{equation}
Here, $f_{Jitter}$ denotes the function of the Jitter operation. This procedure is repeated several times to collect the set of $N_{jitter}$ refined jittered boxes (i.e., $\{b_{i}^{'}\}$). These refined jittered boxes are then fed to the traditional bagging algorithm which helps to obtain optimum refined boxes. Mathematically, this can be stated as:
\begin{equation}
    \hat{b}_{i} = f_{Bagging}(\{b_{i}^{'}\}), 
\end{equation}
where, $f_{Bagging} = max(\cdot)$ is the bagging operation which selects the bounding box candidate with maximum area. The obtained bounding boxes (i.e., $\hat{b}_{i}$) are further passed through an adaptive threshold filter to generate foreground bounding boxes (i.e., $\hat{b}_{i}^{fg}$). Finally, given the pseudo boxes $\mathcal{B}_{reg}$ for training the box regression on unlabeled data, the regression loss is formulated as
\begin{equation}
    L_{pl}^{reg} = \frac{1}{N_b^{fg}}\sum_{i=1}^{N_b^{fg}}l_{reg}(\hat{b}_i^{fg},\mathcal B_{reg}),
\end{equation}
where, $N_b^{fg}$ is the total number of foreground box, $l_{reg}$ is the box regression loss\footnote{We use standard mean absolute error for regression task.}.

\subsection{Update mechanism}
At each iteration, the teacher weights get marginal updates using the student's weights via the update mechanism. The gradually updated teacher is prone to weight fluctuations of the student network when a teacher network mispredicts a label. 
In \cite{xu2021end, tang2021humble}, authors used the Exponential Moving Average (EMA) mechanism to mitigate the negative effect of incorrect pseudo-labels \cite{EMA}.
However, the EMA update mechanism faces an issue of lagging, which limits its performance during sudden weight fluctuations. In this work, we employ EMA's extension, i.e., Double Exponential Moving Average (DEMA) \cite{DEMA2, DEMA1}. The DEMA provides higher weight to the most recent observations and removes the inherent lag as compared to the EMA update mechanism.
Further, the DEMA update mechanism can be mathematically defined as:
\begin{equation}
    w(t)_{ts} = 2 \cdot EMA(t)_{ts} - EMA(EMA(t))_{ts},
\end{equation}
where, the $EMA$ can be expressed as
\begin{equation}
        EMA(t)_{ts} = \alpha^2 w(t)_{ts-1} + (1-\alpha^2) w(s)_{ts}.
\end{equation}
Here, $w(t)_{ts}$ and $w(s)_{ts}$ are the weights of teacher and student network at current timestamp $ts$ and $\alpha$ is set to $0.999$.

\section{Experimental Setup and Discussion}
\subsection{Details of Dataset and Evaluation}
We present our results on benchmark MS-COCO \cite{lin2014microsoft} and Pascal VOC \cite{pascal} datasets.

\noindent \textbf{MS-COCO dataset:} It comprises of more than 118k labeled images (train2017 set), 123k unlabeled images (unlabeled2017 set) and 5k labeled validation dataset (val2017 set). For validation purpose, we follow the principles suggested in \cite{sohn2020simple,tang2021proposal,jeong2019consistency} and the same is discussed below:\vspace{-0.5em}
\begin{itemize}
    \item \textbf{Partially Labeled Data:} Here, 1\%, 5\%, and 10\% of the train2017 set are sampled as labeled data, while remaining unsampled images are operated as unlabeled data. The network is performed on five different folds and evaluated by taking an average of all five folds.\vspace{-0.7em}
    \item \textbf{Fully Labeled Data:} This setting is more challenging. It aims to enhance a trained detector on large-scale labeled data by using the extra unlabeled data. The training process uses the entire train2017 set as the labeled data and unlabeled2017 set as unlabeled dataset.\vspace{-0.5em}
\end{itemize}


\noindent \textbf{Pascal VOC dataset:} 
It takes VOC07 \textit{trainval} set as labeled data having 5,011 images, and 11,540 images from the \textit{trainval} set of VOC12 as an unlabeled data. The performance is evaluated on the test set of VOC07 in three experimental setups: (a) fully supervised on VOC07 labeled set; (b) VOC07 labeled set and VOC12 additional unlabeled set; and (c) VOC07 labeled set and VOC12 \& COCO20cls\footnote{COCO20cls is generated by leaving only COCO images whose object categories overlap with the object categories used in PASCAL VOC07.} as additional unlabeled sets.

    
    
Furthermore, we follow the data augmentation guidelines given in STAC \cite{sohn2020simple} and FixMatch \cite{sohn2020fixmatch} for training and pseudo-label generation. 
For evaluation, we use the mean average precision (mAP)\footnote{Average Precision (AP) is finding the area under precision-recall curve. 
The mAP is defined as the average of AP.} metric with its different variants i.e., mAP at IoU=0.5 (mAP@50), IoU=0.75 (mAP@75) and IoU=0.5:0.95 (mAP). 
\vspace{-0.5em}
\subsection{Training Setups and Hyper-parameter Tuning}
All experiments are performed with NVIDIA A100 dual GPUs. To allow a fair comparison with previous methods, we use Faster R-CNN \cite{ren2015faster} as our default detection network with pre-trained ResNet-50 \cite{he2016deep} as a backbone. For training and inference, 2k and 1k region proposals have been generated using a non-maximum suppression threshold of 0.7. We sample 512 out of 2k proposals as the box candidates in each training step.

The proposed network is trained for 180k with 0.2 data sampling ratio and 720k iterations with 0.5 data sampling ratio for partially labeled data setting and fully labeled data setting, respectively. 
We adopt SGD as optimizer with learning rate of 0.001 divided by 10 at 120k and 160k iterations for partially labeled data setting and at 480k and 680k iterations for fully labeled data setting. Initially, we set foreground threshold to 0.5 and then it adjusts itself adaptively between 0.5 to 0.9. 

We set $N_{jitter}$ to 10 in the proposed Jitter-Bagging module for regression task. The jittered boxes are randomly sampled by adding the offsets on four coordinates, and the offsets are uniformly sampled from [-6\%, +6\%] of the height or width of the pseudo box candidates.


\begin{table}[t!]
\centering
\caption{Comparison with supervised baseline for various \%s of labeled data.} \label{tab:proposed-supervised}
\vspace{-0.5em}
\begin{adjustbox}{width=.47\textwidth,center}
\begin{tabular}{l|cc|cc|cc|cc}
\hline
       & \multicolumn{2}{c|}{\textbf{1\%}} & \multicolumn{2}{c|}{\textbf{5\%}} & \multicolumn{2}{c|}{\textbf{10\%}} & \multicolumn{2}{c}{\textbf{100\%}} \\ \cline{2-9}
       & Supervised  &  Ours    & Supervised    & Ours   & Supervised     & Ours  & Supervised     & Ours    \\ \hline
mAP@50 & 21.3          & 44.6    & 41.0          & 52.1    & 45.1           & 57.6    & 57.6           & 65.2     \\
mAP@75 & 11.0          & 28.0    & 24.4          & 34.6    & 28.4           & 41.0    & 40.4           & 48.1 \\    
mAP & 11.4          & 26.3    & 23.4          & 32.2    & 27.1           & 37.4    & 37.9           & 44.0     \\  \hline
\end{tabular}
\end{adjustbox}
\end{table}
\subsection{Result analysis of proposed network}
\begin{figure}[t!]
    \centering
        \includegraphics[width = 0.49\textwidth, height = 0.5\textheight]{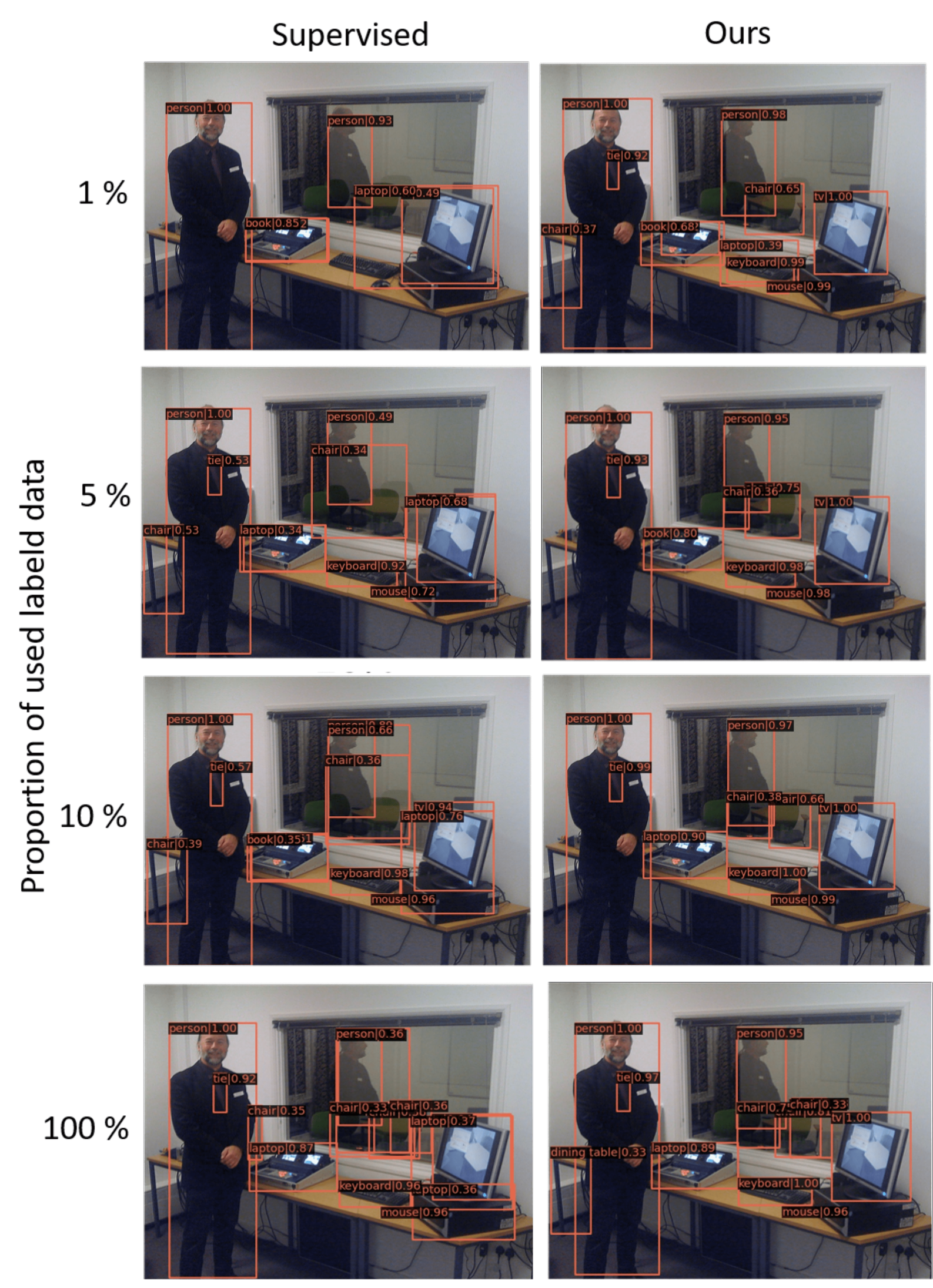}
    \caption{Visual comparison with supervised results.}
    \label{fig:visual1} \vspace{-1em}
\end{figure}

This section provides the result comparison between the proposed network along with its supervised network on MS-COCO dataset. 
The validation performance is tabulated in Table \ref{tab:proposed-supervised}. Here, one can see that our proposed network shows a significant performance improvement than the supervised baseline network in all protocols. In addition, we present the visual comparison for different proportion of labeled data in Figure~\ref{fig:visual1}. Here, it is clearly observed that the proposed network is able to detect tiny and occluded objects with better confidence score than that of the supervised framework.
\vspace{-0.5em}



\subsection{Comparison with state-of-the-art methods}
\subsubsection{Comparison on MS-COCO:}
\textbf{Partially labeled data setting:}
This section compares our network with existing state-of-the-art (SOTA) SSOD methods under the partially labeled data setting. The corresponding average mAP measures of 5 folds are noted in Table \ref{tab:partial}. 
From the table, we can observe that the proposed network outperforms all other methods and obtain +0.25\%, and +1.34\% higher mAP values than previous best performing SOTA methods \cite{zhang2021semi, PseCo} on 1\%, and 10\% labeled data, respectively. While in case of 5\% labeled data setting, it performs slightly inferior than SOTA methods \cite{LabelMatch, PseCo}.


\begin{table}[t!]
\centering
\caption{Comparison between different semi-supervised methods on val2017.}
\label{tab:partial}
\begin{adjustbox}{width=.495\textwidth,center}
\begin{tabular}{l|c|ccc}
\hline
\textbf{Methods}    \quad      & Remarks           & \quad\textbf{1\%} \quad         & \quad \textbf{5\%} \quad   & \quad \textbf{10\%} \quad        \\ \hline
Soft-teacher\cite{xu2021end} &    ICCV 2021          & 20.46 ± 0.39   & 30.74 ± 0.08   & 34.04 ± 0.14   \\
ACRST \cite{zhang2021semi} & Arxiv 2021 & 26.07 ± 0.46 & 31.35 ± 0.13 & 34.92 ± 0.22 \\
DDT \cite{DDT} & AAAI 2022 & 19.44 ± 0.32 & 29.92 ± 0.12 & 33.46 ± 0.18 \\
CST \cite{CST} & ACM MM 2022 & 22.73 ± 0.14 & 30.83 ± 0.08 & 33.90 ± 0.17 \\
Active Teacher \cite{ActiveTeacher} & CVPR 2022 & 22.20 & 30.07 & 32.58 \\
MA-GCP \cite{MAGCP} & CVPR 2022 & 21.31 ± 0.28 & 31.67 ± 0.16 & 35.02 ± 0.26 \\
DCST \cite{DCST} & IJCAI 2022 & 23.02 ± 0.23 & 32.10 ± 0.15 & 35.20 ± 0.20 \\
PseCo \cite{PseCo} & ECCV 2022 & 22.43 ± 0.36 & 32.50 ± 0.08 & 36.06 ± 0.24 \\
LabelMatch \cite{LabelMatch} & CVPR 2022 & 25.81 ± 0.28 & \textbf{32.70 ± 0.18} & 35.49 ± 0.17 \\
Polishing Teacher \cite{PoshingTeacher} & AAAI 2023 & 23.55 ± 0.25 & 32.10 ± 0.15 & 35.30 ± 0.15 \\
Ours & --- & \textbf{26.32 ± 0.35}   & 32.21 ± 0.08   & \textbf{37.40 ± 0.15}   \\ \hline
\end{tabular}
\end{adjustbox}
\end{table}
\begin{table}[t!]
\centering
\caption{Comparison with other state-of-the-arts under the setting of fully labeled data of train2017 set. }
\label{tab:supervisedF}
\begin{adjustbox}{width=.475\textwidth,center}
\begin{tabular}{l|c|cc}
\hline
\textbf{Methods}    &   Remarks          & \quad \textbf{Extra Dataset} \quad      & \quad \textbf{mAP} \quad      \\ \hline
Self-training \cite{zoph2020rethinking}   &  NIPS 2020     & ImageNet+OpenImages & 41.1 $\xrightarrow{+0.8}$ 41.9 \\
Soft-teacher \cite{xu2021end}   & ICCV 2021         & unlabeled2017       & 40.9 $\xrightarrow{+3.6}$ 44.5 \\
MA-GCP \cite{MAGCP}    & CVPR 2022        & unlabeled2017       & 40.9 $\xrightarrow{+5.0}$ 45.9 \\
LabelMatch \cite{LabelMatch}  & CVPR 2022          & unlabeled2017       & 40.3 $\xrightarrow{+5.0}$ 45.3 \\
DDT \cite{DDT}    & AAAI 2022        & unlabeled2017       & 37.6 $\xrightarrow{+4.6}$ 42.2 \\
CST \cite{CST}    & ACM MM 2022        & unlabeled2017       & 37.6 $\xrightarrow{+5.7}$ 43.3 \\
DTG-SSOD \cite{DTG-SSOD}  & Arxiv 2022          & unlabeled2017       & 40.9 $\xrightarrow{+4.8}$ 45.7 \\
PseCo \cite{PseCo}    & ECCV 2022        & unlabeled2017       & 41.0 $\xrightarrow{+5.1}$ 46.1 \\
DCST \cite{DCST}     & IJCAI 2022       & unlabeled2017       & 40.9 $\xrightarrow{+3.7}$ 44.6 \\
Ours  & ---   & unlabeled2017       & 37.9 $\xrightarrow{\textbf{+6.1}}$ 44.0 \\ \hline
\end{tabular}
\end{adjustbox}
\end{table}
\noindent\textbf{Fully labeled data setting:}
Here, we compare our network with other methods in the fully labeled data setting. Since the reported performance of the supervised baseline varies, we report the results of the comparison methods and their baseline simultaneously in Table \ref{tab:supervisedF}. The additional unlabeled dataset required to improve the baseline performance is also mentioned here and the corresponding improvement is noted in this table. One can see that the proposed network shows a larger performance gain (i.e., +6.1\%) than the existing state-of-the-art methods.

\begin{table}[t]
\centering
\caption{Comparison with other state-of-the-art methods on the VOC07 test set.}
\label{tab:PascalVOC}
\begin{adjustbox}{width=.47\textwidth,center}
\begin{tabular}{l|c|ccc}
\hline
\textbf{Model} & Remarks & \textbf{mAP} & \textbf{mAP@50} & \textbf{mAP@75} \\ \hline \hline
\multicolumn{1}{l|}{\begin{tabular}[l]{@{}c@{}}VOC07 labeled data \\ (Supervised)\end{tabular}} & --- & 41.91       & 66.0          & 45.1          \\ \hline \hline
\multicolumn{5}{l}{VOC07 labeled set + VOC12 unlabeled set}    \\ \hline
RPL \cite{li2021rethinking} & Arxiv 2021 & 54.60 & 79.00   & 59.40  \\
CST \cite{CST} & ACM MM 2022 & 51.50 & 78.70   & ---  \\
MA-GCP \cite{MAGCP} & CVPR 2022 & --- & 81.72   & ---  \\
DDT \cite{DDT} & AAAI 2022 & 54.70 & 82.40   & 59.80  \\
LableMatch \cite{LabelMatch} & CVPR 2022 & 55.11 & \textbf{85.48}   & ---  \\
Polishing Teacher \cite{PoshingTeacher} & AAAI 2023 & 52.40 & 82.50   & ---  \\
Ours  & ---  & \textbf{56.92} & 82.04 & \textbf{62.84}  \\ \hline \hline
\multicolumn{5}{l}{VOC07 labeled set + VOC12 \& COCO20cls unlabeled set} \\ \hline
Unbiased teacher \cite{liu2021unbiased}  & ICLR 2021  & 50.34     & 78.82   & ---  \\
Instant Teaching \cite{zhou2021instant}  & CVPR 2021 & 50.80     & 79.90   & 55.70  \\
RPL \cite{li2021rethinking}  & Arxiv 2021  & 56.10 & 79.60   & 61.20   \\
CST \cite{CST}  & ACM MM 2022  & 53.50 & 80.50   & ---   \\
DDT \cite{DDT}  & AAAI 2022  & 55.90 & \textbf{82.50}   & 61.10   \\
Ours & ---  & \textbf{57.10} & 82.21   & \textbf{63.47}     \\ \hline
\end{tabular}\end{adjustbox}
\end{table}

\subsubsection{Comparison on Pascal VOC:}
We also evaluate our network on the Pascal VOC benchmark dataset and the comparison is presented in Table \ref{tab:PascalVOC}. When utilizing VOC07 labeled and VOC12 unlabeled data, the proposed network obtains +15.01\%, +16.04\% and +17.74\% higher values compared to the supervised setting on mAP, mAP@50 and mAP@75, respectively.
The proposed network also outperforms the previous best performing SOTA methods \cite{LabelMatch, DDT} by +1.81\%, and +3.04\% on mAP, and mAP@75, respectively. To analyze how increasing unlabeled data can help to improve performance, we have used the COCO20cls dataset as an additional unlabeled set. As a result, the proposed network shows an absolute improvement of +15.19\%, +16.21\% and +18.37\% compared to the fully supervised baseline. We can also see that the proposed network outperforms the SOTA methods \cite{li2021rethinking} by +1.00\%, and +2.27\% on mAP, and mAP@75, respectively. These results verify that our network can further improve object detection by using more unlabeled data.


\subsection{Ablation Analysis}\label{sec:abl}
All experiments are carried out on 10\% partially labeled setting of MS-COCO. However, the analysis on 1\% and 5\% data settings are covered in the supplementary material.

\textbf{Effectiveness of adaptive threshold filter:} 
To prove the effectiveness of proposed adaptive threshold filter, few experiments have been carried out and 
the corresponding results are presented in Table \ref{Tab:threshold}. Here, we can see that the proposed adaptive threshold filter performs better than the static threshold values. 
To check its effectiveness over the threshold module proposed by Li \etal \cite{li2021rethinking}, we employed their thresholding module in our framework. The corresponding results are added in Table \ref{Tab:threshold}, which is marginally inferior to the proposed thresholding module. 
In our adaptive mechanism, we have used discrete thresholding to reduce fluctuations in the threshold value, Further, we trained a variant of our model with a continuous form of threshold and found lower performance than proposed discrete form.
\begin{table}[t!]
\caption{Analysis to validate adaptive threshold mechanism.}\label{Tab:threshold}
\begin{adjustbox}{width=.99\linewidth,center}
\begin{tabular}{l|ccc}
\hline
     \textbf{Proposed Network}    & \textbf{mAP} & \textbf{mAP@50} & \textbf{mAP@75} \\ \hline
 with static 0.7 threshold      & 33.0  & 52.7   & 36.8   \\
with static 0.8 threshold      & 36.3  & 55.6   & 39.1   \\
with static 0.9 threshold      & 37.0  & 56.4   & 40.3   \\
with dynamic thresholding \cite{li2021rethinking}  & 37.1  & 56.8   & 40.6    \\ \hline
with continuous form-based threshold & 36.1 & 55.9   & 39.5   \\
with proposed adaptive threshold & \textbf{37.4}  & \textbf{57.5}   & \textbf{41.0}  \\ \hline
\end{tabular}
\end{adjustbox}
\end{table}

\begin{table}[t!]
\centering
\caption{
Analysis to validate the Jitter-Bagging module.} \label{Tab:JitterBagging}\vspace{-0.5em}
\begin{adjustbox}{width=.8\linewidth,center}
\begin{tabular}{l|ccc}
\hline
\multicolumn{1}{l|}{\textbf{Proposed Network}} & \textbf{mAP}  & \textbf{mAP@50} & \textbf{mAP@75} \\ \hline
without Jitter-Bagging & 36.2 & 56.9   & 39.7   \\
with Box Jittering \cite{xu2021end} & 36.8 & 57.0 & 40.1   \\
with Jitter-Bagging  & \textbf{37.4} & \textbf{57.5}   & \textbf{41.0}  \\ \hline
\end{tabular}
\end{adjustbox}
\end{table}
\textbf{Importance of Jitter-Bagging module:} The ablation analysis of Jitter-Bagging module is presented in Table \ref{Tab:JitterBagging}, where we can see that the proposed Jitter-Bagging module achieves highest performance and shows +1.2\% absolute improvement in mAP measure over without Jitter-Bagging module. Interestingly, when we employ the Box Jittering \cite{xu2021end} in our network, we observe that the proposed Jitter-Bagging still obtains +0.6\% higher mAP than Box Jittering.

\begin{table}[t!]
\centering
\caption{Analysis to validate introduced losses for classification.} \label{abl:loss}\vspace{-0.5em}
\begin{adjustbox}{width=.99\linewidth,center}
\begin{tabular}{l|ccc}
\cline{1-4}
\textbf{Network}                                              &\, \textbf{mAP}  & \, \textbf{mAP@50} &\, \textbf{mAP@75} \\ \hline 
Case I:\qquad $L^{cls}_{fg} + L^{cls}_{bg}$                    & 36.8 & 56.9   & 40.4   \\
Case II: \quad\ $L^{cls}_{fg} + L^{cls}_{bg} + L_{bg}^{sim}$               & 37.2 & 57.3   & 40.6   \\
Case III: \quad $L^{cls}_{fg} + L^{cls}_{bg} + L_{fg-bg}^{dissim}$ & 37.1 & 57.2   & 40.6   \\
Proposed: \, $L^{cls}_{fg} + L^{cls}_{bg} + L_{bg}^{sim} + L_{fg-bg}^{dissim}$                                       & \textbf{37.4} & \textbf{57.5}   & \textbf{41.0}   \\ \hline 
\end{tabular}
\end{adjustbox}
\end{table}
\begin{figure}[t!]
    \centering
    \includegraphics[width = 0.99\linewidth]{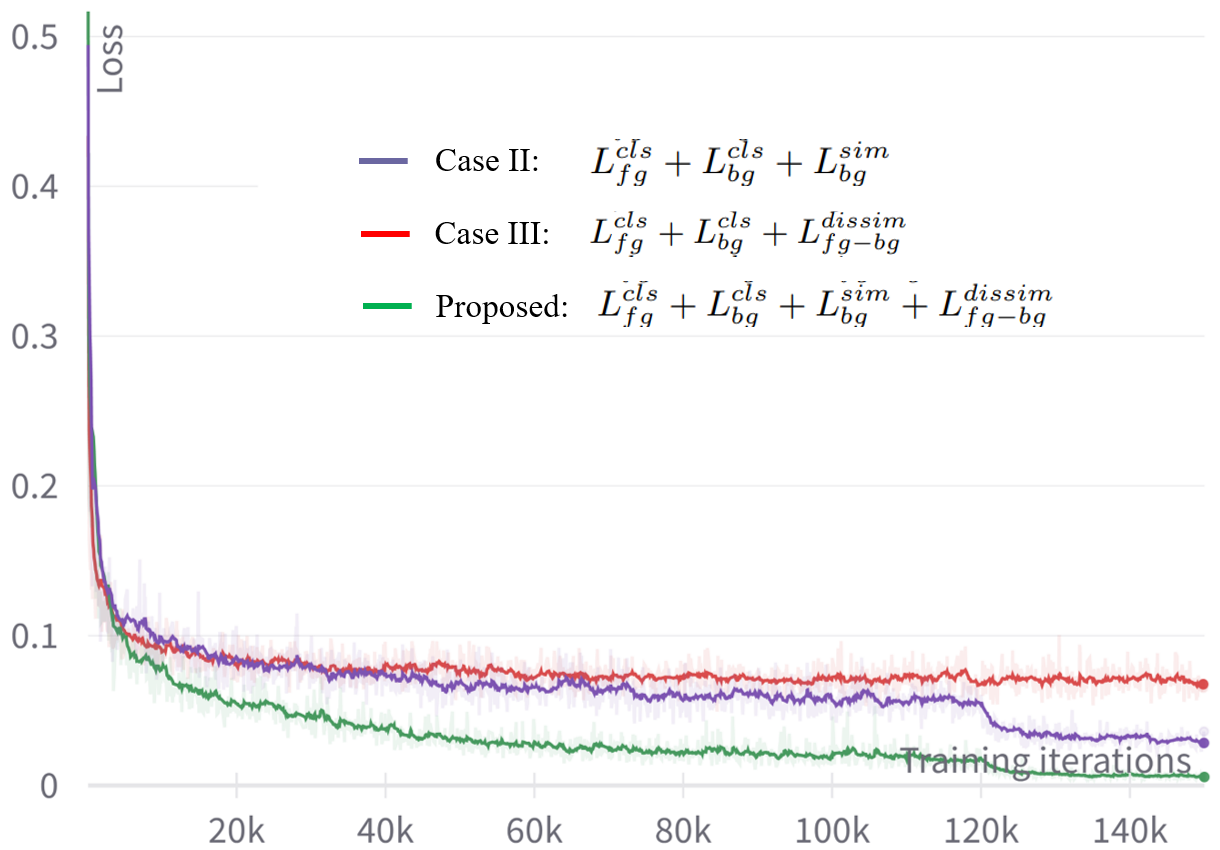}
    \vspace{-0.5em}
    \caption{Analysis of the introduced classification loss during training iterations. }
    \label{fig:loss-training}
\end{figure}
\textbf{Effect of losses for classification:} For classification task, we introduce two novel losses; background similarity loss i.e., $L_{bg}^{sim}$ and foreground-background dissimilarity loss i.e., $L_{fg-bg}^{dissim}$. To check its effectiveness, the proposed network is trained without both introduced losses (i.e., Case I), with background similarity loss (i.e., Case II) and with foreground-background dissimilarity loss (i.e., Case III). The corresponding measures are depicted in Table \ref{abl:loss}. Here, it can be noticed that both introduced losses help the proposed network to obtain better mAP measures. Additionally, Figure \ref{fig:loss-training} shows the effect of loss values during the training iteration. Here, it can be seen that the proposed network with both losses converges better than others. 


\begin{table}[t!]
\centering
\caption{Analysis to validate the DEMA update mechanism.} \label{abl:updatemechanism}
\vspace{-0.5em}
\begin{adjustbox}{width=.7\linewidth,center}
\begin{tabular}{l|ccc}
\cline{1-4}
\textbf{Network}  &\, \textbf{mAP}  & \, \textbf{mAP@50} &\, \textbf{mAP@75} \\ \hline 
Deep Copy           & 32.1 & 51.5   & 33.8   \\
EMA update    & 36.2 & 56.4   & 39.9   \\
DEMA update & \textbf{37.4} & \textbf{57.5}   & \textbf{41.0}  \\ \hline
\end{tabular}
\end{adjustbox}
\end{table}
\textbf{Effect of update mechanism:} To verify the effectiveness of the DEMA update mechanism, we ablate the proposed network trained using EMA mechanism as well as employing deep copy configuration (i.e., weights of teacher network are copied from student network). The corresponding results are noted in Table \ref{abl:updatemechanism}, where it can be seen that DEMA helps to obtain +1.2\% higher mAP measure, demonstrating its efficacy over EMA update mechanism.

\begin{table}[t!]
\centering
\caption{Analysis to check importance of label generators.}\label{abl:labelgenerator}
\vspace{-0.5em}
\begin{adjustbox}{width=.99\linewidth,center}
\begin{tabular}{l|ccc}
\cline{1-4}
\textbf{Network}                                              &\, \textbf{mAP}  & \textbf{mAP@50} & \textbf{mAP@75} \\ \hline 
Proposed (Both label generators)  & \textbf{37.4} & \textbf{57.5}   & \textbf{41.0}   \\ 
\quad - w/o label generator (weak augmented data)   & 35.2 & 55.4   & 38.9   \\
\quad - w/o label generator (strong augmented data) & 36.6 & 56.8   & 40.4   \\
\hline
\end{tabular}
\end{adjustbox}
\end{table}
\textbf{Importance of label generator module:} In our proposed network, we use two label generator modules; one associated with weak augmented samples while the other is based on strong augmented sample. To see the importance of this setting, the proposed network with individual label generator is trained and the obtained results are presented in Table \ref{abl:labelgenerator}. Here, it is observed that the proposed network with both label generators outperforms the individual label generator settings.


\section{Conclusion}
In this paper, we present an end-to-end teacher-student network to address the class imbalance issue in semi-supervised object detection. It successfully examines the effect of class imbalance on pseudo-label generation and proposes novel learning mechanisms to improve the pseudo-label quality. Specifically, we tackle the high false-negative and low precision rates using the proposed adaptive threshold mechanism and refine optimal bounding boxes using our Jitter-Bagging module.
We further introduce two novel losses based on background and foreground bounding boxes to improve the pseudo-label recall rate so that it can detect small objects as foreground. Finally, our extensive experimentation shows that the proposed network outperforms existing state-of-the-art SSOD methods on MS-COCO and Pascal VOC benchmark datasets.



\balance
{\small
\bibliographystyle{ieee_fullname}
\bibliography{egbib.bib}
}

\end{document}